# Aging display's effect on interpretation of digital pathology slides

Ali R. N. Avanaki[a], Kathryn S. Espig[a], Sameer Sawhney[c], Liron Pantanowitz[c], Anil V. Parwani[c], Albert Xthona[a], Tom R. L. Kimpe[b]
[a]Barco Healthcare, Beaverton, OR; [b]Barco Healthcare, Kortrijk, Belgium;
[c]Department of Pathology, University of Pittsburgh Medical Center, Pittsburgh, PA

## ABSTRACT

It is our conjecture that the variability of colors in a pathology image effects the interpretation of pathology cases, whether it is diagnostic accuracy, diagnostic confidence, or workflow efficiency. In this paper, digital pathology images are analyzed to quantify the perceived difference in color that occurs due to display aging, in particular a change in the maximum luminance, white point, and color gamut. The digital pathology images studied include diagnostically important features, such as the conspicuity of nuclei. Three different display aging models are applied to images: aging of luminance & chrominance, aging of chrominance only, and a stabilized luminance & chrominance (i.e., no aging). These display models and images are then used to compare conspicuity of nuclei using CIE $\Delta E_{2000}$, a perceptual color difference metric. The effect of display aging using these display models and images is further analyzed through a human reader study designed to quantify the effects from a clinical perspective. Results from our reader study indicate significant impact of aged displays on workflow as well as diagnosis as follow. As compared to the originals (no-aging), slides with the effect of aging simulated were significantly more difficult to read (p-value of 0.0005) and took longer to score (p-value of 0.02). Moreover, luminance+chrominance aging significantly reduced inter-session percent agreement of diagnostic scores (p-value of 0.0418).

## 1. INTRODUCTION AND BACKGROUND

It is our conjecture that the variability of colors in a pathology image effects the interpretation of pathology cases, whether it is diagnostic accuracy, diagnostic confidence, or workflow efficiency. In this paper, digital pathology images are analyzed to quantify the perceived difference in color that occurs due to display aging, in particular a change in the maximum luminance, white point, and color gamut. The digital pathology images studied include diagnostically important features, such as the conspicuity of nuclei.

Three different display aging models are applied to images: aging of luminance & chrominance, aging of chrominance only, and a stabilized luminance & chrominance (i.e., no aging). These display models and images are then used to compare conspicuity of nuclei using CIE $\Delta E_{2000}$, a perceptual color difference metric.

The effect of display aging using these display models and images is further analyzed through a human reader study protocol designed to study the effects from a clinical perspective.

It is common that with age, a display's maximum luminance will decrease and its chrominance will change. There are several reasons why displays degrade over time. Backlights use phosphors to convert electricity to light and these phosphors wear out over time and use [7]. Optical materials (many layers) are partially transparent and become more absorptive as they age [6, 11]. Liquid crystal chemicals break down, also becoming more absorptive and less responsive to pixel drive [5]. Electrical contacts become corroded, increasing resistance, and further reducing drive levels [5, 10]. Dust and dirt accumulate in the display, and thus block more light. For all of these reasons stated, the amount of light produced by a display decreases over time. This has the consequence of reducing the maximum display luminance ($L_{max}$).

All of the degradations occur at different rates for different light frequencies. This results in color shifts and a change in the color gamut of the display in terms color points (i.e. xy from CIE xyY color space). In addition, all of the degradations occur at different rates for screen location. This results in uniformity changes within the display.

## 2. METHODS

### 2.1 Models of aging

We consider the following models of display aging in this paper.
- Aging luminance and chrominance: Consumer displays generally follow this aging model, and show signs of degrading luminance and chrominance over time. The causes are discussed in the background section.

- Stabilized luminance with aging chrominance: Lmax is kept constant over time and the chrominance changes over time. The Barco MDCC 6130 display behaves this way due the use of a continuous feedback loop to a light sensor integrated inside the display.

- Stable luminance and chrominance (negligible aging): Lmax is kept constant. Chrominance is also kept constant due to color stabilization. The Barco MDCC 6230 display follows this model.

### 2.2 Display aging simulation

In order to determine how a display, and ultimately an image, will look after a certain amount of time, we model two properties: its change in maximum luminance and change in chrominance. We modeled the change in chrominance by first measuring xyY values for the white point and primary tips (red, green, blue) 31 times over 18000 hours. These measurements were taken with Barco display MDCC 6130, which verified the expectation that luminance was stable and chrominance varied over time. To model the behavior of display at a certain age, we use the standard sRGB behavior [4] altered to match these measurements. More specifically, first a piece-wise linear model is used to interpolate the measurements in time. Then, at a given instance of time, another piece-wise linear model is used to find what happens to an arbitrary color (e.g., a pixel of an image), given the known (from time-interpolated measurements) changes to white point and the primaries. Finally, the aged color is expressed in terms of standard sRGB so that it can be rendered on a standard (non-aged) sRGB display. The latter procedure for modeling the effect of aging on an image destined for a standard sRGB display is described in the following.

1. XYZ values are calculated for each measurement [3] using the following formulas.

$$X = \frac{xY}{y}, Y = Y, and\ Z = (1 - x - y)Y/y$$

2. Matrixes T'$_i$, i = w, r, g, & b (white, red, green, and blue), are calculated. Calculation of T'$_r$ is shown below. Other matrixes are calculated similarly.

$$T'_r = \begin{pmatrix} X_w & X_g & X_b \\ Y_w & Y_g & Y_b \\ Z_w & Z_g & Z_b \end{pmatrix} \begin{pmatrix} 1 & 0 & 0 \\ 1 & 1 & 0 \\ 1 & 0 & 1 \end{pmatrix}^{-1}$$

where $(X_i, Y_i, Z_i)$, i = w, r, g, & b are calculated from display measurements for white and primary tips in Step 1. To model luminance aging, $Y_i$ values should be reduced simulating loss of luminance.

3. Transform matrixes T$_i$, i = w, r, g, & b are given by

$$T_i = \frac{1}{3} \sum_{j \neq i} T'_j$$

Based on the RGB values of the pixels being processed (see Step 5), one of $T_i$ matrixes is used (instead of RGB-to-XYZ conversion matrix used in standard sRGB [8]) to calculate the aging-modeled XYZ values from RGB.

4. Next, the sRGB standard nonlinearity [8] is applied to each pixel (r'', g'', b'') of the image:

$$r' = \begin{cases} \dfrac{r''}{12.92}, & r'' \leq 0.04045 \\ \left(\dfrac{a+r''}{a+1}\right)^{2.4}, & r'' > 0.04045 \end{cases} \quad \text{with } a = 0.055$$

values of g' and b' are calculated similarly.

5. For the transform below, $T_i$ is selected if (r', g', b') is closest to color point i = w, r, g, or b (white or one of the primary tip). For example, $T_b$ is selected to transform (0.02, 0.012, 0.99).

$$\begin{pmatrix} X \\ Y \\ Z \end{pmatrix} = T_i \begin{pmatrix} r' \\ g' \\ b' \end{pmatrix}$$

This completes the modeling of aging in XYZ space.

6. The forward transform given in [4] is used to map the image back to the standard sRGB. Therefore, the original and the aged image (i.e., after simulated aging) may be shown side by side on the same sRGB-compliant display.

## 2.3 Reader study

We conducted a reader study to determine if using non-aged and aged images have an effect on pathologist workflow and diagnosis. The images selected exemplify a diverse group of cases that show nuclear and architectural heterogeneity. We included a variety of images for this study, including: Surgical pathology formalin fixed tissue (H&E stain), frozen section tissue (H&E stain), cytology liquid-based specimens (ThinPrep, Pap stain), and cytology direct smears (Diff Quik stain). Figure 1 shows a subset of the images used.

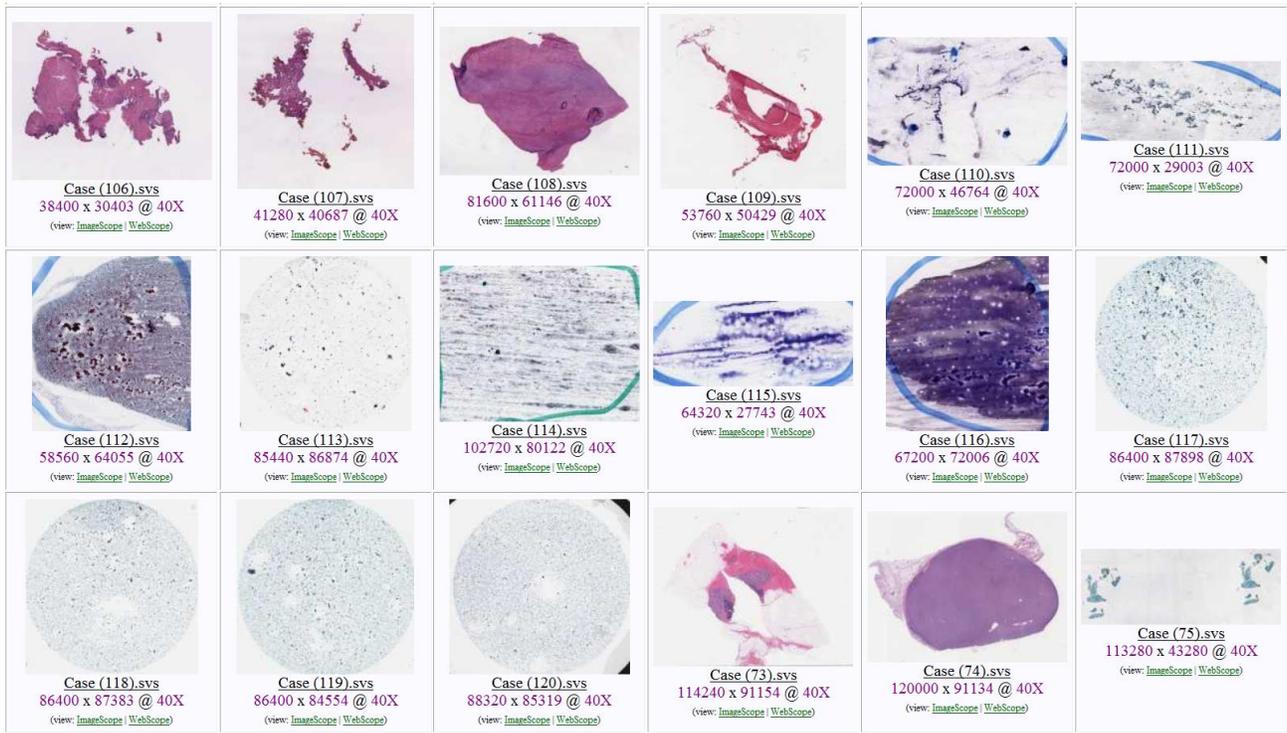

**Figure 1: A** selection of the images viewed in the reader study. All images are shown non-aged.

Additionally, the select cases provide a variable amount of cytological characteristics that pathologists identify on a daily basis during sign out. These features are integral during histologcal analysis and are necessary to be represented on a digital slide to validate its usage.

All images were viewed on the Barco display MDCC 6230. The following protocol was used:

- One pathologist selected a set of 120 images with diagnostically significant features.

- In the first reading session, a second pathologist scores the 120 images, 40 with chroma aging, 40 with chroma+luma aging, and 40 non-aged. The images were presented to the pathologist in random order. The pathologist recorded three items for each image.

    1. Time To Score (TTS)
    2. Ease of Reading (EOR; score from 1 to 10, larger means easier to read)
    3. Diagnosis: negative (0 or 1+) or positive (2+ or 3+).

- The second reading session is conducted after a washout period of 24 days. The second pathologist scores the 120 images, all non-aged, in randomized order.

To study the impact of aging on workflow, average and standard deviation of TTS and EOR for the three conditions (non-aged, with chroma aging, and with chroma+luma aging) were calculated and inspected for statistically significant differences.

To investigate the impact of aging on diagnosis, the percent agreement (PA) between scores given in session 1 and session 2, for each type of aging (i.e., PA between scores given to slides that were not aged in session 1 and the same slides in session 2, PA between scores given to chroma-aged slides in session 1 and the same slides in session 2, and PA between scores given to chroma+luma-aged slides in session 1 and the same slides in session 2) is calculated.

## 3. RESULTS

### 3.1 Simulation results

Three exemplary images are shown in Figures 2-4, the non-aged image at the center with the aged versions around it as follows. From top, clockwise: Chroma aged only at 10,000 hours, chroma aged only at 18,000 hours, chroma and luminance aging 10,000 hours, and chroma and luminance aging 18,000 hours. We assumed luminance to be 67% and 40% of the original at 10,000 hours and 18,000 hours respectively.

Figure 5 shows the aging in terms of the color difference between aged and non-aged pixels of Figure 2, average over all pixels, with 95% confidence intervals (i.e., ± twice standard deviation). The color difference is measured in CIE ΔE2000.

The expected lifetime of displays, based on different CIE ΔE2000 tolerances, is listed in Table 1. It lists the lifetime based on the four different criteria: Max diff under 3 ΔE2000, Avg diff under 3 ΔE2000, Avg diff under 5 ΔE2000, Max diff under 5 ΔE2000.

The color difference between non-aged and aged images are shown in Figure 6. The top image shows the difference between non-aged and 8100 hours of chroma-only aging. The bottom images shows the difference between non-aged and 4100 hours of chroma+luma aging images. The scale on the right goes from 1 to 8 CIE ΔE2000.

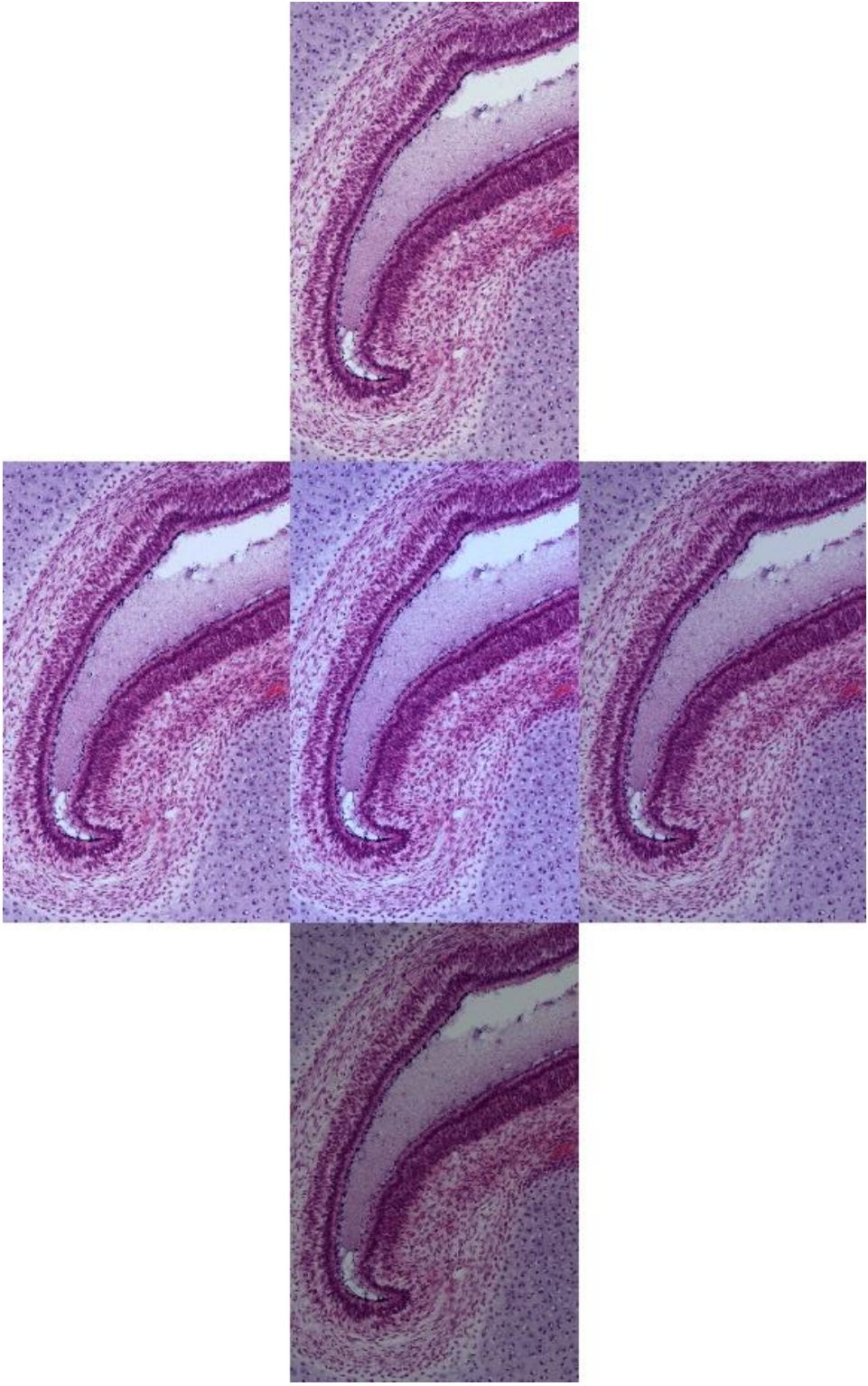

**Figure 2.** Example 1 of display aging effect. Original (non-aged) image is at the center with the aged versions around it as follows. From top, clockwise: Chroma aged only at 10,000 hours, chroma aged only at 18,000 hours, chroma and luminance aging 10,000 hours, and chroma and luminance aging 18,000 hours.

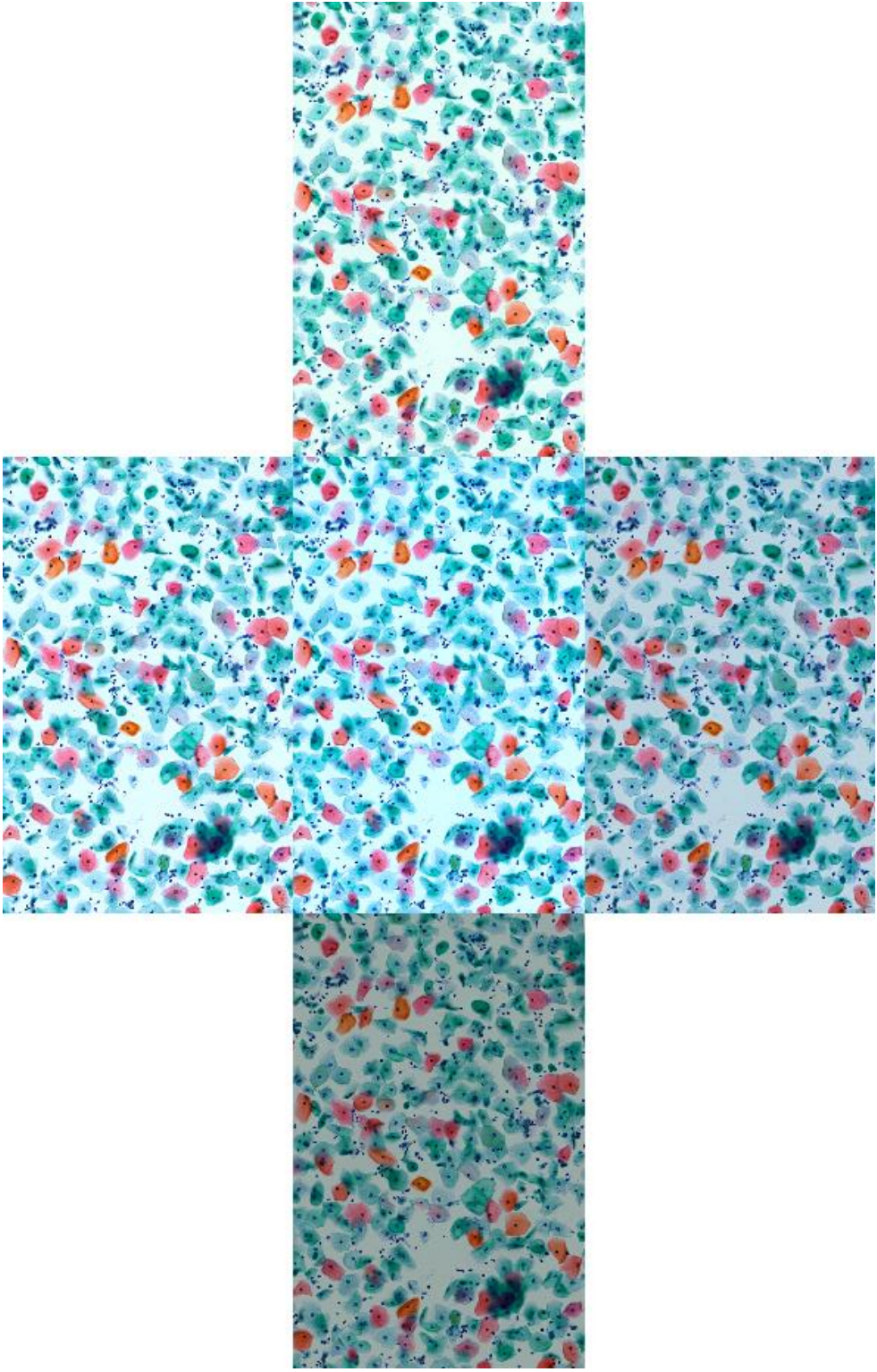

**Figure 3.** Example 2 of display aging effect. Original (non-aged) image is at the center with the aged versions around it as follows. From top, clockwise: Chroma aged only at 10,000 hours, chroma aged only at 18,000 hours, chroma and luminance aging 10,000 hours, and chroma and luminance aging 18,000 hours.

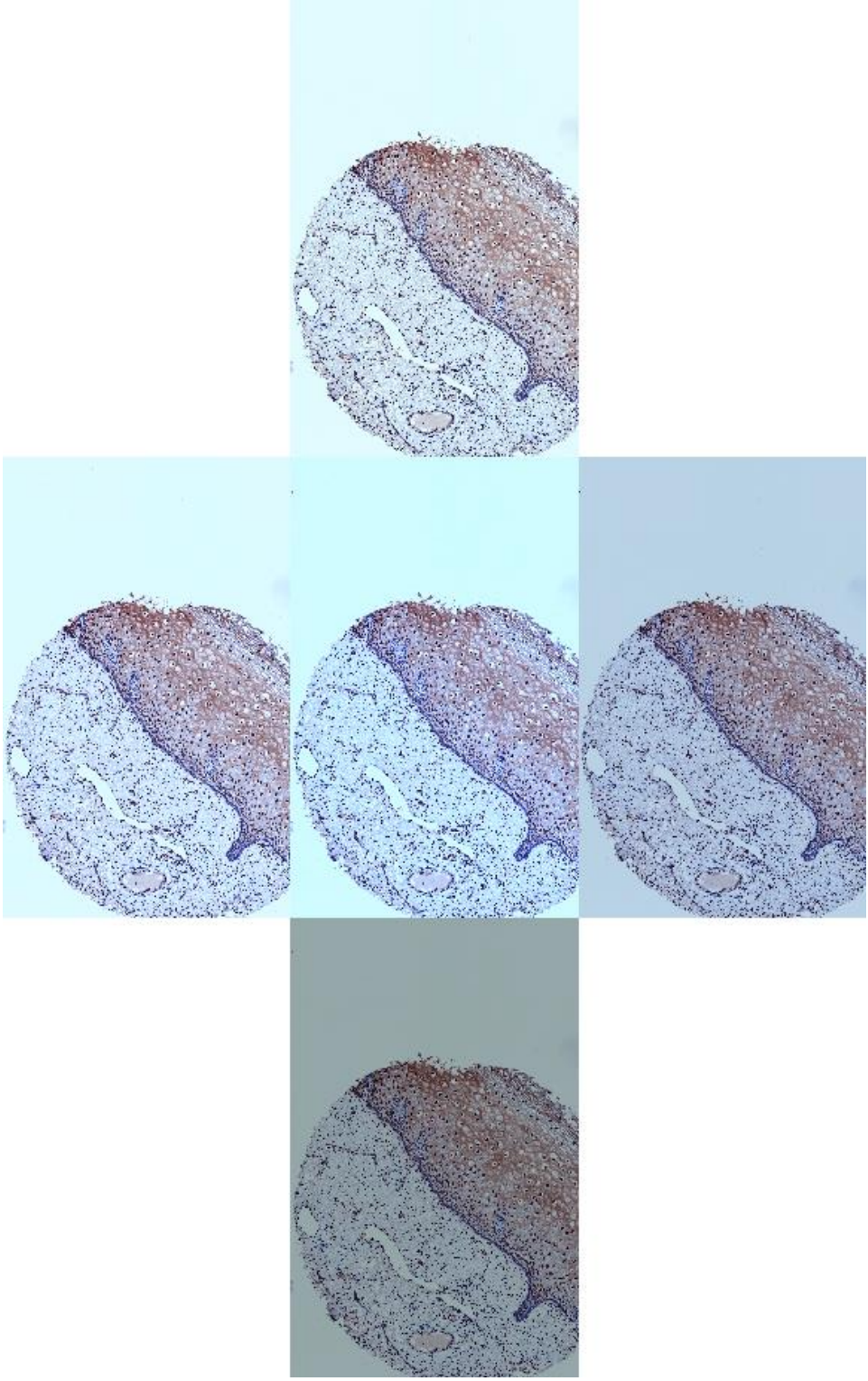

**Figure 4.** Example 3 of display aging effect. Original (non-aged) image is at the center with the aged versions around it as follows. From top, clockwise: Chroma aged only at 10,000 hours, chroma aged only at 18,000 hours, chroma and luminance aging 10,000 hours, and chroma and luminance aging 18,000 hours.

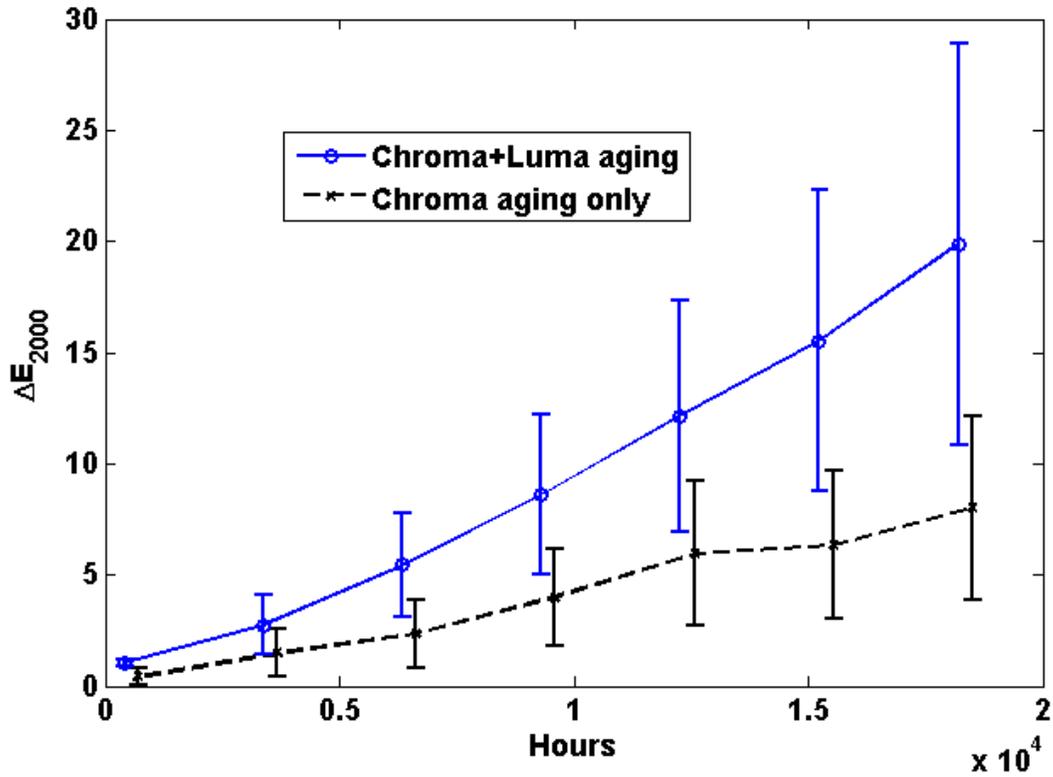

**Figure 5.** Effect of aging in terms of color difference between aged and non-aged versions of image shown in Figure 2, averaged over all pixels, with 95% confidence intervals as error bars.

**Table 1.** Life time, in hours, considering various tolerances in color difference with un-aged display, measured from curves in Figure 5 (i.e., based on Figure 2 appearances).

|  | Max diff under 3 $\Delta E_{2000}$ | Avg diff under 3 $\Delta E_{2000}$ | Max diff under 5 $\Delta E_{2000}$ | Avg diff under 5 $\Delta E_{2000}$ |
|---|---|---|---|---|
| Chroma aging only | 4500 | 7800 | 8100 | 11100 |
| Chroma+Luma aging | 2200 | 3700 | 4100 | 5800 |

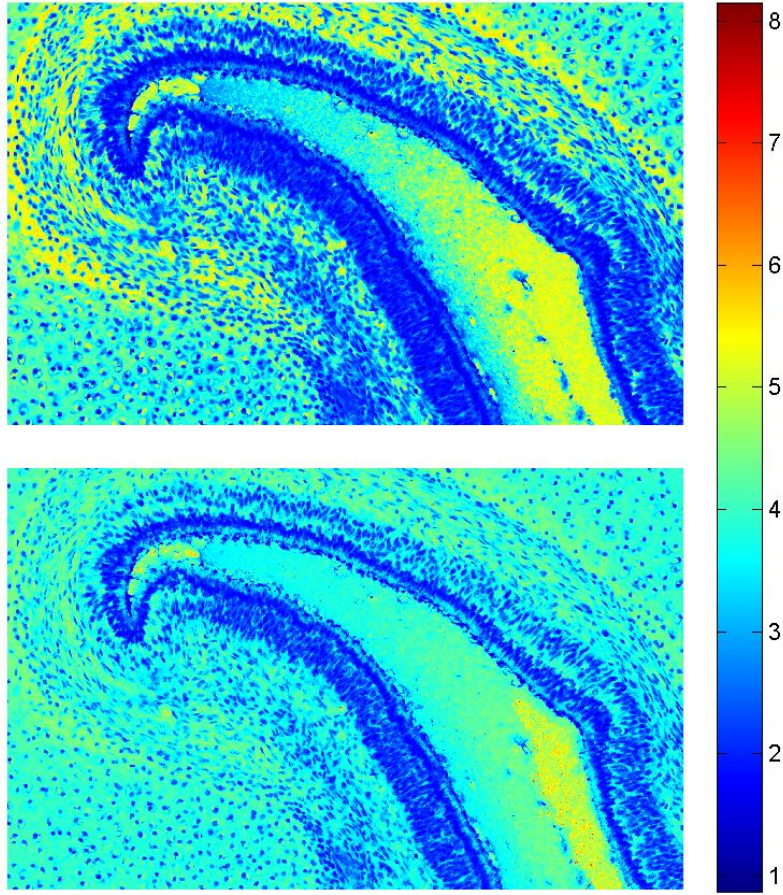

**Figure 6.** Color difference map between un-aged and aged. The bar on the right gives a visual indication of the difference in CIE $\Delta E_{2000}$. Top: 8100 hours of chroma-only aging Bottom: 4100 hours of chroma+luma aging images.

### 3.2 Reader study results

**Effect of aging on workflow**

Average of time to score (TTS) and ease of reading (EOR) data for different types of aging, are provided in Table 2, along with 95% confidence intervals.

**Table 2.** Average time to score (TTS) and Ease of reading (EOR), along with their 95% confidence intervals, for three types of aging.

| Type of aging | Time to score (sec) | Ease of reading (1: hard, 10: easy) |
|---|---|---|
| Non-aged | 41.0 ± 5.4 | 9.7 ± 0.2 |
| Chroma aging | 50.1 ± 7.3 | 8.7 ± 0.5 |
| Chroma + luma aging | 51.2 ± 6.4 | 8.4 ± 0.4 |

Two-sample (unpaired) t-tests [9] at $\alpha = 0.05$ significance level were used to check if the mean TTS and EOR for different types of aging are different. For EOR, the null hypothesis is rejected (i.e., significant mean difference) in comparing non-aged vs chroma-aged (p-value of 0.0005) and non-aged vs chroma+luma-aged (p-value of 2e-7). For TTS, the null hypothesis is rejected in comparing non-aged vs chroma+luma-aged (p-value of 0.02). (Note that for comparison of non-aged vs chroma-aged, in terms of mean TTS, p-value is 0.054).

There is a notable difference between TTS and EOR of non-aged and aged images. TTS for non-aged images is about 10 seconds shorter. EOR for non-aged images is about 1 point higher, i.e., non-aged images are notably easier to read. Based on this table, however, there is not a significant difference between chroma-aged and chroma+luma aged images in terms of EOR and TTS.

**Effect of aging on diagnostics**

Percent agreement (PA) for each type of aging between the scores given in session 1 and session 2 is calculated. The 95% confidence intervals for PA values are calculated using Wilson's method [12]. The results are given in Table 3.

**Table 3.** Percent agreement between scores given in reading sessions 1 and 2 for different types of display aging.

| Aging (in session 1 only) | # of cases scores agreed | # of cases scores disagreed | Percent agreement |
|---|---|---|---|
| None (no-aging) | 27 | 12 | 69.23<br>95% CI: [53.6  81.4] |
| Chroma-only | 21 | 19 | 52.50<br>95% CI: [37.5  67.1] |
| Chroma+luma | 20 | 20 | 50.00<br>95% CI: [35.2 64.8] |

As compared to the value for non-aged slides, the inter-session PA is dropped about 20% for chroma-only and chroma+luma types of aging. The PA for chroma-only and chroma+luma types of aging are rather close (52.5 vs 50%). This would suggest that it is mostly the change in chrominance that results into reduced PA.

When PA formulated as an average, we can perform a t-test to check if PA differences are significant. To formulate PA as an average, we make a new population composed of 0s, when scores are differ between sessions, and 1s, when scores agree between sessions. The average of such population is PA. Results of unpaired one-tailed t-tests at $\alpha = 0.05$ significance level assuming equal variances (checked by F-tests) are reported in Table 4. Use of one-tailed tests is justified since we are certain luminance or chrominance stability cannot worsen diagnosis [13].

**Table 4.** Significance of PA differences. PA values are listed Table 3.

| Null hypothesis | One-tailed t-test result (p-value) |
|---|---|
| $PA_{no-aging} <= PA_{chroma\ aging}$ | Null hypothesis cannot be rejected (0.0656) |
| $PA_{no-aging} <= PA_{chroma+luma\ aging}$ | Null hypothesis is rejected (0.0418) |
| $PA_{chroma\ aging} <= PA_{chroma+luma\ aging}$ | Null hypothesis cannot be rejected (0.413) |

Based on Table 4, there is a significant difference between no-aging and chroma+luma aging inter-session PA values.

## 4. DISCUSSION AND CONCLUSION

It is clear from visual inspection of the images that display aging strongly affects the visualized color of the pathology images. The CIE $\Delta E_{2000}$ numbers further confirm the difference. It is notable that even during typical lifetime of a display system, the CIE $\Delta E_{2000}$ reach values that are very high. For example: for typical usage pattern, even during the first year of use a chroma+luma aged display can reach average changes of more than 3 CIE $\Delta E_{2000}$ units (with maximum values being even higher).

The reader study also confirms that display aging affects the workflow. Pathologists consider it faster and easier to read non-aged images. Inter-session percent agreement (PA) of diagnostic scores for non-aged slides is about 20% higher than that of aged slides, though PA confidence intervals overlap due to the size of the study. Nevertheless the different between no-aging PA and chroma+luma aging PA is statistically significant. This is a very important finding since it suggests that display aging could negatively influence diagnostic quality. We plan to add a third pathologist (as the second reader) to the study to further increase the statistical strength of the study.

Technology is available to compensate for (both luma and chroma) aging of display systems. The results of this paper suggest that integrating such technology in display systems for digital pathology could reduce reading time, increase ease of reading, and improve diagnostic quality.

## ACKNOWLEDGEMENT

Ali Avanaki wishes to thank Bastian Piepers for help in derivation of display aging model.